\documentclass[10pt,journal,compsoc]{IEEEtran}

\usepackage{soul}
\usepackage{url}
\usepackage{ulem}
\usepackage{booktabs}
\usepackage{amsmath}
\usepackage{amsthm} 
\usepackage{amsfonts}  
\usepackage{amssymb}
\usepackage{multicol}
\usepackage{multirow}
\usepackage{graphicx}
\usepackage{booktabs}
\usepackage{colortbl}
\usepackage{stfloats}
\usepackage{comment}
\usepackage{makecell}
\usepackage[noend]{algpseudocode}
\usepackage{algorithmicx,algorithm}
\usepackage{pifont}
\usepackage{color}
\usepackage{microtype}
\usepackage{cleveref}
\usepackage[]{footmisc}

\usepackage{soul}
\usepackage{url}
\usepackage{ulem}
\usepackage{booktabs}
\usepackage{amsmath}
\usepackage{amsthm} 
\usepackage{amsfonts}  
\usepackage{amssymb}
\usepackage{multicol}
\usepackage{multirow}
\usepackage{graphicx}
\usepackage{booktabs}
\usepackage{colortbl}
\usepackage{stfloats}
\usepackage{comment}
\usepackage{makecell}
\usepackage[noend]{algpseudocode}
\usepackage{algorithmicx,algorithm}
\usepackage{pifont}
\usepackage{color}
\usepackage{microtype}

\ifCLASSOPTIONcompsoc
  \usepackage[nocompress]{cite}
\else
  \usepackage{cite}
\fi

\begin{document}

\title{HC$^2$L: Hybrid and Cooperative Contrastive Learning for Cross-lingual Spoken Language Understanding
}

\author{Bowen~Xing and~Ivor~W.~Tsang,~\IEEEmembership{Fellow,~IEEE} 
\IEEEcompsocitemizethanks{\IEEEcompsocthanksitem Bowen Xing is with Beijing Key Laboratory of Knowledge Engineering for Materials Science, School of Computer and Communication Engineering, University of Science and Technology Beijing.\protect\\
E-mail: bwxing714@gmail.com
\IEEEcompsocthanksitem Ivor Tsang is with CFAR, Agency for Science, Technology and Research;
       IHPC, Agency for Science, Technology and Research; 
       School of Computer Science and Engineering, Nanyang Technological University;
       Australian Artificial Intelligence Institute, University of Technology Sydney. \protect\\
E-mail: ivor\_tsang@cfar.a-star.edu.sg}
}

\markboth{Journal of \LaTeX\ Class Files,~Vol.~xx, No.~x, xx~2024}%
{Shell \MakeLowercase{\textit{et al.}}: Bare Demo of IEEEtran.cls for Computer Society Journals}

\IEEEtitleabstractindextext{%
\begin{abstract}
State-of-the-art model for zero-shot cross-lingual spoken language understanding performs \textit{cross-lingual unsupervised contrastive learning} to 
achieve the label-agnostic semantic alignment between each utterance and its code-switched data.
However, it ignores the precious intent/slot labels, whose label information is promising to help capture the label-aware semantics structure and then leverage supervised contrastive learning to improve both source and target languages' semantics.
In this paper, we propose Hybrid and Cooperative Contrastive Learning to address this problem.
Apart from cross-lingual unsupervised contrastive learning, we design a holistic approach that exploits \textit{source language supervised contrastive learning}, \textit{cross-lingual supervised contrastive learning} and \textit{multilingual supervised contrastive learning} to perform label-aware semantics alignments in a comprehensive manner. Each kind of supervised contrastive learning mechanism includes both single-task and joint-task scenarios.
In our model, one contrastive learning mechanism's input is enhanced by others. Thus the total four contrastive learning mechanisms are cooperative to learn more consistent and discriminative representations in the virtuous cycle during the training process.
Experiments show that our model obtains consistent improvements over 9 languages, achieving new state-of-the-art performance.

\end{abstract} 
}

\maketitle

\IEEEdisplaynontitleabstractindextext

\IEEEpeerreviewmaketitle

\section{Introduction}\label{sec:introduction}
Spoken language understanding (SLU) plays a crucial role in task-oriented dialog systems \cite{slu,idsf,glgin,coguiding}.
It includes two subtasks: intent detection, which is a sentence-level classification task, and slot filling, which is a sequence labeling task.
Great progress has been achieved in the past decade, while current SLU methods require a large amount of data, which is impractical in some scenarios.
To this end, zero-shot cross-lingual spoken language understanding \cite{conneau2019cross,its2sptr,ars2sptr,gl-clef} has been explored and attracted increasing interest because it can significantly reduce the effort for data annotation and transfer the task knowledge learned from the high-resource language into the target low-resource language.

\begin{figure}[t]
 \centering
 \includegraphics[width = 0.49\textwidth]{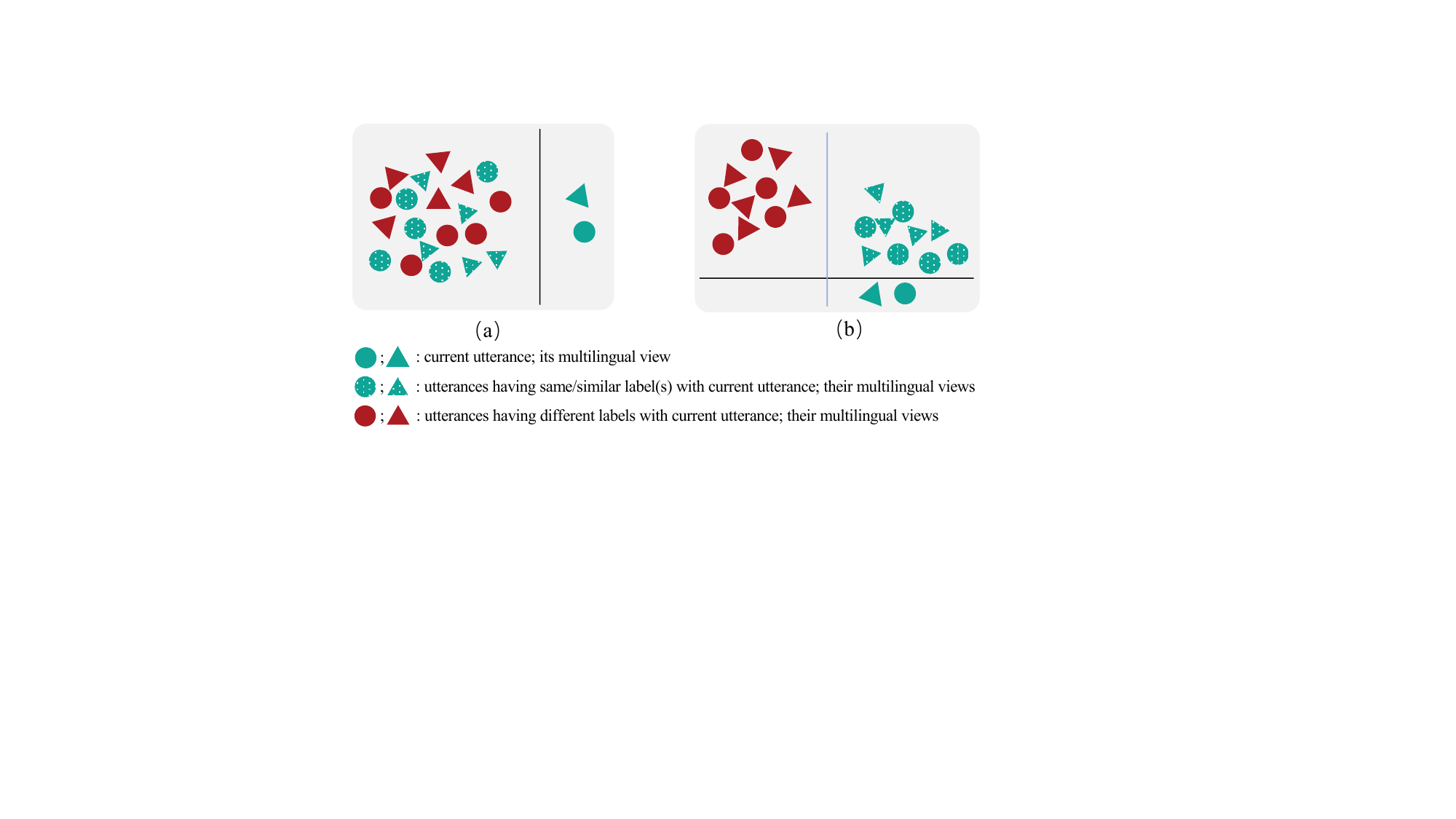}
 \caption{Conceptual comparison of the contrastive learning mechanisms of GL-CLEF (a) and our HC$^2$L (b). The black line denotes the margin caused by the cross-lingual unsupervised contrastive learning proposed by GL-CLEF. The blue line denotes the margin caused by our proposed three kinds of supervised contrastive learning mechanisms.}
 \label{fig: concept}
\end{figure}

Most previous models can only conduct implicit multilingual semantics alignment via parameters sharing \cite{attention-informed,slot-align-and-rec,cosda-ml,crossing-the-conversational-chasm}.
GL-CLEF \cite{gl-clef}, which is the up-to-date state-of-the-art model, proposes to leverage cross-lingual unsupervised contrastive learning (CL) to perform explicit semantics alignment between the utterance and its multilingual view obtained by code-switching \cite{cosda-ml}.
Its contrastive learning mechanism is conceptually illustrated in Fig. \ref{fig: concept} (a).
GL-CLEF pulls together the current utterance and its multilingual view in the semantic space.
At the same time, the current utterance is pushed apart from all other utterances as well as their multilingual views.
Despite the significant improvements that GL-CLEF achieves, we find that it suffers from a drawback: its contrastive learning mechanism does not consider any label information.
This leads to two issues.
First, some utterances and multilingual views have the same/similar label(s) as the current utterance. Thus, it is intuitive to pull them together.
However, they are toughly separated only because they are not the current utterance's multilingual view.
Second, some utterances and multilingual views have different labels, so their semantics are supposed to be pushed apart.
However, GL-CLEF cannot achieve this due to ignoring the label information.

\begin{figure}[t]
 \centering
 \includegraphics[width = 0.49\textwidth]{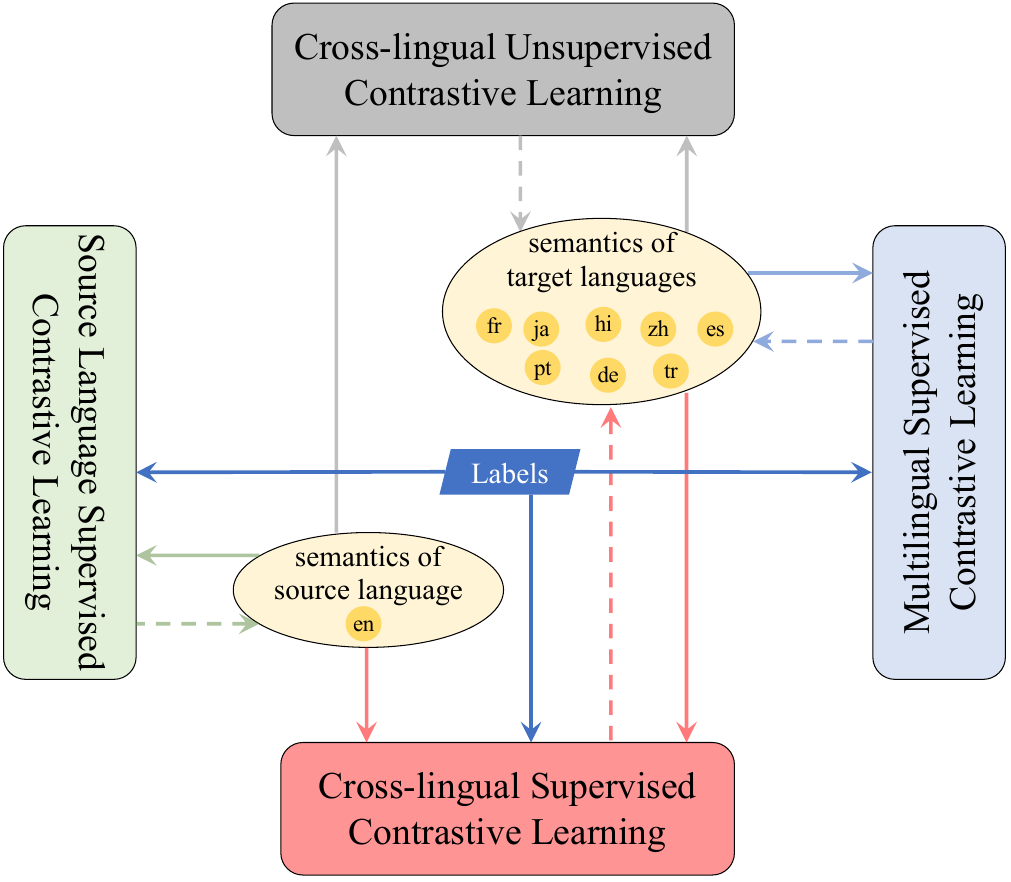}
 \caption{Illustration of the cooperation of the hybrid contrastive learning mechanisms in our HC$^2$L model. Solid arrows denote input. Dashed arrows denote enhancement.} 
 \label{fig: interaction}
\end{figure}

\begin{table*}[t]
\centering
\fontsize{8}{11}\selectfont
\caption{Summary of the four contrastive learning (CL) mechanisms in our model. }
\setlength{\tabcolsep}{2.5mm}{
\begin{tabular}{l|l|l}
\midrule
 \textbf{CL Mechanisms} & \textbf{Input}  & \textbf{Enhancement} \\ \midrule
Cross-lingual Unsupervised CL  & Source Language Semantics; Multilingual View Semantics & Multilingual View Semantics\\ \hline
Source Language Supervised CL  & Labels; Source Language Semantics & Source Language Semantics\\ \hline
Cross-lingual Supervised CL & Labels; Source Language Semantics; Multilingual View Semantics & Multilingual View Semantics\\ \hline
Multilingual Supervised CL & Labels; Multilingual View Semantics & Multilingual View Semantics\\
  \midrule
\end{tabular}}
\label{table: summary cl}
\end{table*}

Therefore, we argue that the precious and sufficient label information is urgent to be leveraged to perform supervised contrastive learning to capture the label-aware semantics structure.
And we discover three promising perspectives for designing supervised contrastive learning: source language, cross-lingual and multilingual.
Then we propose HC$^2$L: Hybrid and Cooperative Contrastive Learning to address the above problems, as conceptually illustrated in Fig.\ref{fig: concept}(b).
We design three kinds of supervised contrastive learning mechanisms: (1) \textit{\textbf{source language supervised contrastive learning}}, which enhances the source language semantics via label-aware semantics alignment; (2) \textit{\textbf{cross-lingual supervised contrastive learning}}, which transfers the knowledge from the source language to target languages via performing the label-aware alignment between the source language semantics and the multilingual view semantics; (3) \textit{\textbf{multilingual supervised contrastive learning}}, which aligns the multilingual view semantics that has the same/similar label(s).
Each of the three kinds of supervised contrastive learning comprehensively includes both of the single-task (intent/slot) and the joint-task (intent+slot) supervised contrastive learning mechanisms.
The intent/slot supervised contrastive learning mechanisms leverage intent/slot labels to capture high-level semantic structure via performing label-aware semantics alignment.
However, there is no given label for the joint task.
To this end, we construct it by ourselves using the given intent and slot labels, and we propose the joint-task multi-label supervised contrastive learning, which can model the dual-task correlations.

To achieve explicit cross-lingual semantics alignment, HC$^2$L also includes the cross-lingual unsupervised contrastive learning proposed by GL-CLEF \cite{gl-clef}.
As shown in Fig. \ref{fig: interaction}, the total four kinds of contrastive learning mechanisms in our model have mutual influence and interdependencies: the input semantics of one contrastive learning mechanism is reinforced by other contrastive learning mechanisms.
In this way, the hybrid contrastive learning mechanisms in our model can cooperate with each other to learn better and better semantic representations in the training procedure.

Thanks to the proposed supervised contrastive learning mechanism, our HC$^2$L model have two advantages over previous models:
\begin{itemize}
\item It can learn more \textit{consistent} semantic representations across different languages.
\item It can learn more \textit{discriminative} semantics representations across different classes.
\end{itemize}

We evaluate our model on MultiATIS++, which is a benchmark including 9 different languages. Experiment results show that our model obtains about 10\% average improvements over the previous best-performing model on overall accuracy.
Further analysis verifies the effectiveness of our proposed contrastive learning mechanisms and it is proven that our proposed supervised contrastive learning mechanisms contribute more than the cross-lingual unsupervised contrastive learning.
The visualizations of learned representations show that our model can learned more consistent representations across different languages. And in the same time, our model effectively pushes apart the semantics corresponding to different classes.

\section{Related Works}\label{sec: relatedwork}


\subsection{Zero-shot Cross-lingual Spoken Language Understanding}
Spoken language understanding \cite{yao2014spoken,zhang2016joint,vu2016bi,su2018time} is a core component of dialog systems.
It usually includes two subtasks: intent detection and slot filling.
Intent detection is a sentence-level classification task aiming to predict the intent expressed in an utterance.
Slot filling is a sequence labeling task that assigns a slot label to each word.
In recent years, as researchers have realized the correlations between intent detection and slot filling,
a group of models are proposed to jointly tackle the two tasks via leveraging the correlative information  \cite{liuandlane,slot-gated,selfgate,sfid,cmnet,qin2019,agif,jointcap,slotrefine,glgin,coguiding,relanet,darerpami,coguiding-pami}.
Co-guiding Net \cite{coguiding} makes the first time to model the mutual guidance between the multi-intent detection and slot filling via heterogeneous semantics-label graphs.
ReLa-Net \cite{relanet} exploits the dual-dependencies and leverages them for dual-task interaction and joint decoding.

However, these SLU models largely rely on rich-source training data, which is not always practicable, especially for some low-source languages.
To solve this problem, zero-shot cross-lingual SLU is explored and has attracted increasing attention.
Since mBERT \cite{bert} is a strong baseline for cross-lingual language understanding, some methods are proposed to improve mBERT at the pre-training stage \cite{conneau2019cross,unicoder,xlm-r,yang2020alternating,xue-etal-2021-mt5,chi-etal-2021-infoxlm}.

Besides, some other works aim to perform semantics alignment between the source language and target languages at the fine-tuning stage.
Attention-informed mixed-language training \cite{attention-informed} proposes code-mixing to construct multilingual training samples containing phrases from both of source and target languages.
 And CoSDA \cite{cosda-ml} further proposes multilingual code-switching to better perform multilingual semantics alignment. 
GL-CLEF \cite{gl-clef}, which is the up-to-date state-of-the-art, proposes to perform explicit multilingual semantics alignment via cross-lingual unsupervised CL.
LAJ-MCL \cite{multilevelCL} proposes to model the utterance-slot-word structure by a multi-level contrastive learning framework.
FC-MTLF \cite{FC-MTLF} proposes to leverage the neural machine translation task to improve cross-lingual SLU.
DiffSLU \cite{diffslu} leverages a powerful diffusion model to enhance the mutual guidance between the slot and intent.
Differently, we make the first attempt to perform three kinds of label-aware semantics alignments via source language, cross-lingual and multilingual supervised CL.


\subsection{Contrastive Learning for NLP}
Contrastive learning has been widely adopted to improve representations in NLP tasks \cite{zhang-etal-2021-pairwise,knn-multilabelcl,knn-intentcl,wang2022incorporating}.
In the natural language inference task,  pairwise supervised CL \cite{zhang-etal-2021-pairwise} bridges semantic entailment and contradiction understanding with high-level categorical concept encoding.
Hierarchy-guided contrastive learning \cite{wang2022incorporating} directly incorporates the hierarchy into the text encoder for hierarchical text classification.
In this paper, we propose three kinds of supervised CL (e.g., source language supervised CL, cross-lingual supervised CL and multilingual supervised CL) to achieve comprehensive label-aware semantics alignments for zero-shot cross-lingual spoken language understanding .


\section{HC$^2$L} 

\begin{figure*}[t]
 \centering
 \includegraphics[width = \textwidth]{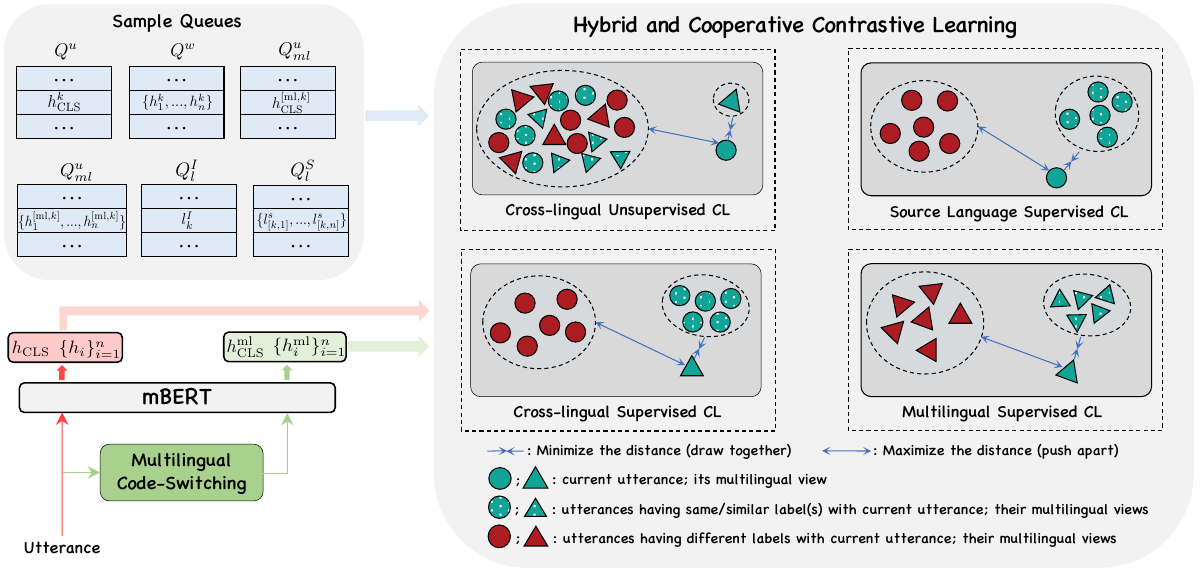}
 \caption{The architecture of our HC$^2$L model. For simplicity, the SLU prediction module is omitted, and our three supervised contrastive learning mechanisms are illustrated in single-label manner.}
 \label{fig: model}
\end{figure*}

Zero-shot cross-lingual SLU aims to train the model in a source language (e.g., English) and then directly apply it to target languages (e.g., French, Japan, Chinese) for testing.
And there are two subtasks:
\begin{itemize}
\item Intent detection. It is a sentence-level classification problem aiming to predict the correct intent label $l^I$.
\item Slot filling. It is a token-level sequence labeling task mapping the input utterance word sequence $X = \{x_1, ..., x_n\}$ to the slot sequence $\{l^s_1, ..., l^s_n\}$, where $n$ denotes the word number.
\end{itemize}
In this section, we introduce our HC$^2$L model in detail.
Its architecture is shown in Fig. \ref{fig: model}, and its four contrastive learning mechanisms are summarized in Table \ref{table: summary cl}.

\subsection{Backbone Framework}
\subsubsection{Encoder}
Following state-of-the-art method \cite{gl-clef}, we adopt the pre-trained mBERT model to encode the input utterance word sequence, and the representation of the first sub-token of a word is used for the word representation.
Then we obtain the word hidden states:
\begin{equation}
 \mathbf{H} = \{h_\texttt{CLS}, h_1, ..., h_n\}
 \end{equation}
 where \texttt{[cls]} is the special token at the beginning of the input sequence and $h_\texttt{[CLS]}$ is taken as the sentence representation; $h_t$ denotes the first sub-token representation of word $x_t$.
Then we follow \cite{gl-clef} to generate the multilingual code-switched data
(multilingual view), which is fed to Multilingual BERT (mBERT) \cite{bert} to generate the hidden states:
\begin{equation}
 \mathbf{H}^{\text{ml}} = \{h^{\text{ml}}_{\texttt{CLS}}, h_1^{\text{ml}}, ..., h_n^{\text{ml}}\}.
\end{equation}
\subsubsection{Intent Detection Decoder}.
The sentence representation $h^{\text{ml}}_{\text{CLS}}$ is fed to a softmax classifier to predict the intent label:
\begin{equation}
l^{I} = \operatorname{softmax}(W_I \ h^{\text{ml}}_{\texttt{CLS}} + b_I)
\end{equation}
where $W_I$ and $b_I$ denote weight matrix and bias.

\subsubsection{Slot Filling Decoder}.
For word $x_t$, we feed its representation $h^{\text{ml}}_{t}$ into the slot classifier to predict its slot label:
\begin{equation}
l^{s}_t = \operatorname{softmax}(W_S \ h^{\text{ml}}_{t} + b_S)
\end{equation}
where $W_S$ and $b_S$ denote weight matrix and bias.

\subsection{Sample Queues}
Since our model performs both unsupervised and supervised contrastive learning, 
inspired by \cite{moco}, we maintain a set of sample queues that store not only the previously encoded features but also their labels. \\
1. Utterance representation queue: $Q^u = \{h^k_{\texttt{CLS}}\}^K_{k=1}$. 
It stores the sentence representations of previously encoded source language utterances, where $K$ is the length of the queue.\\
2. Word representation queue: $Q^w=\{h^k_1, ..., h^k_n\}^K_{k=1}$.
It stores the word representation sequence of previously encoded source language utterances.\\
3. Multilingual view utterance representation queue: $Q^u_{ml}=\{h^{[\text{ml},k]}_{\texttt{CLS}}\}^K_{k=1}$. 
It stores the sentence representations  of previously encoded code-switched utterances (multilingual views).\\
4. Multilingual view word representation queue: $Q^w_{ml} = \{h^{[\text{ml},k]}_1, ..., h^{[\text{ml},k]}_n\}^K_{k=1}$.  
It stores the word representation sequence  of previously encoded code-switched utterances (multilingual views).\\
5. Intent label queue: $Q^I_l=\{\hat{l}^I_k\}^K_{k=1}$.
It stores the one-hot intent labels  of previously encoded utterances.\\
6. Slot label queue: $Q^S_{l}=\{\hat{l}^s_{[k,1]}, ..., \hat{l}^s_{[k,n]}\}^K_{k=1}$
It stores the one-hot slot label sequence  of previously encoded utterances.


The queues are updated with the current batch's features and labels while dequeuing the oldest ones. 
And each utterance and its multilingual view share the same intent label and slot labels. 
For instance, $h^k_{\texttt{CLS}}$ and $h^{[\text{ml},k]}_{\texttt{CLS}}$ correspond to the same one-hot intent label $\hat{l}^I_k$;
$h^k_1$ and $h^{[\text{ml},k]}_1$ correspond to the same one-hot slot label $\hat{l}^s_{[k,1]}$.

\subsection{Cross-lingual Unsupervised Contrastive Learning}
Our model includes cross-lingual unsupervised contrastive learning, whose effectiveness has been verified in \cite{gl-clef}.
The sentence and word representations of the current utterance's multilingual view are positive samples, while all other representations in $Q^u$, $Q^w$, $Q^u_{ml}$ and $Q^w_{ml}$ are negative samples.\\
\textbf{Intent}. The cross-lingual intent unsupervised contrastive learning mechanism aims to align the sentence representations of the current utterance and its multilingual view.
Specifically, it can be formulated as follows:
\begin{equation}
\mathcal{L}^{I}_{un} = -\operatorname{log}\frac{e^{\frac{h_{\texttt{CLS}} \cdot h^{\text{ml}}_{\texttt{CLS}}} {\tau} }}{e^{\frac{h_{\texttt{CLS}}\cdot h^{{\text{ml}}}_{\texttt{CLS}}}{\tau}} +\sum_k^K{[e^{\frac{h_{\texttt{CLS}}\cdot h_{\texttt{CLS}}^{k}}{\tau}} + e^{\frac{h_{\texttt{CLS}}\cdot h^{[\text{ml},k]}_{\texttt{CLS}}}{\tau}}]}}  
\end{equation}
where $\tau$ denotes the temperature.\\
\textbf{Slot}. As slot filling is token-level, the cross-lingual slot unsupervised contrastive learning mechanism aims to perform token-level semantics alignment.
Specifically, it can be formulated as follows:
\begin{equation}
\mathcal{L}_{un}^{S} = -\frac{1}{n^2}\sum^n_{i}\sum^n_{j}{\operatorname{log}\frac{e^{\frac{h_{i} \cdot h^{\text{ml}}_{j}} {\tau} }}{e^{\frac{h_{i}\cdot h^{{\text{ml}}}_{j}}{\tau}} + \sum_k^K{[e^{\frac{h_{i}\cdot h_{j}^{k}}{\tau}} + e^{\frac{h_{i}\cdot h^{[\text{ml},k]}_{j}}{\tau}}}]}} 
\end{equation}
\textbf{Intent-Slot}. The cross-lingual global intent-slot unsupervised contrastive learning mechanism aims to model the global semantic alignment for both intent and slot.
Specifically, it can be formulated as follows:
\begin{equation}
\begin{split}
\mathcal{L}_{\text{un}}^{\text{GIS}} &= -\frac{1}{n}\sum^n_{j}\operatorname{log}\frac{e^{\frac{h_{\texttt{CLS}} \cdot h_{j}} {\tau} }}{e^{\frac{h_{\texttt{CLS}}, h_{j}}{\tau}} + \sum_k^K{[e^{\frac{h_{\texttt{CLS}}, h_{j}^{k}}{\tau}}  +  e^{\frac{h_{\texttt{CLS}}, h^{[\text{ml},k]}_{j}}{\tau}}]}} + \\
 &-\frac{1}{n}\sum^n_{j}\operatorname{log}\frac{e^{\frac{h_{\texttt{CLS}} \cdot h^{{\text{ml}}}_{j}} {\tau} }}{e^{\frac{h_{\texttt{CLS}}, h^{{\text{ml}}}_{j}}{\tau}}  +  \sum_k^K{[e^{\frac{h_{\texttt{CLS}}, h_{j}^{k}}{\tau}}  +  e^{\frac{h_{\texttt{CLS}}, h^{[\text{ml},k]}_{j}}{\tau}}]}} 
\end{split}
\end{equation}

The motivation behind this design is that in a single sentence, its slots and intent are usually highly related from the semantics perspective. Therefore, GL-CLEF takes the intent representation ($h_{\texttt{CLS}}$) in a sentence and its own slots' representations ($h_j$) to naturally constitute a form of positive pairs, while the slots' representations of other sentences can form negative pairs.

\subsection{Source Language Supervised contrastive learning}
Regarding the current utterance as the anchor, we propose the source language supervised CL to perform source language label-aware semantics alignment.
It draws together the anchor's sentence (word) representation and source language samples in $Q^u$ ($Q^w$) or pushes them apart regarding whether they have the same/similar label(s).

\subsubsection{Intent Supervised contrastive learning}
Taking the current utterance's sentence representation and the samples in $Q^u$ as input,
source language intent supervised contrastive learning pulls together the anchor's sentence representation and the ones of positive samples, while distinguishing the anchor's sentence representation from the ones of negative samples.
And we adopt the cosine function with temperature $\tau'$ to measure the similarity of the two representations:
\begin{equation}
s(a,b) = \frac{a^{T}\cdot b}{\|a\|\cdot \|b\| \cdot \tau'}.
\end{equation}

The samples sharing the same intent label with the anchor is regarded as positive samples, while other ones are negative samples.
However, the positive samples and negative samples are mixed together in $Q^u$. 
To this end, we propose to use the Hadamard product of the current utterance's one-hot intent label and the queue sample's one to automatically retrieve the positive samples.
Denoting current utterance as $i$, this contrastive learning mechanism can be formulated as follows:
\begin{equation}
\begin{split}
\mathcal{L}_{\text{slscl}}^{\text{I}} &= -\sum^{K}_k \frac{\mu_{ik}}{\sum_j^K \mu_{ij}} \operatorname{log}\frac{{ e^{s(h^i_{\text{CLS}}, h^k_{\text{CLS}})}}}{\sum^{K}_j{e^{s(h^i_{\text{CLS}}, h^j_{\text{CLS}})}}}\\
\mu_{ik} &= \hat{l}^I_i\odot \hat{l}^I_k
\end{split}
\end{equation}
where $\mu_{ik}$ equals 0 or 1, indicating whether the $k$-th sample in $Q^u$ is a positive sample; $\odot$ denotes Hadamard product. 

\subsubsection{Slot Supervised Contrastive Learning}
It aims to align the source language's word representations that have the same slot label.
Similarly, we adopt the Hadamard product of one-hot slot labels to automatically retrieve positive word representation samples from $Q^w$.
This contrastive learning mechanism can be formulated as follows:
\begin{equation}
\begin{split}
\mathcal{L}_{\text{slscl}}^{\text{S}} &= -\frac{1}{n^2}\sum^n_{i}\sum^n_{j} \sum^{K}_k  \frac{\mu_{i}^{[k,j]}}{\sum_a^K\mu_{ia}} \operatorname{log}\frac{ e^{s(h_i, h^k_j)}}{\sum^{K}_a{e^{s(h_i, h^a_j)}}}\\
\mu_{i}^{[k,j]} &= \hat{l}^s_i \odot \hat{l}^s_{[k,j]}
\end{split}
\end{equation}
where $\mu_{i}^{[k,j]}$ equals 0 or 1, indicating whether the $j$-th word representation of the $k$-th sample in $Q^w$ is a positive sample of the current utterance's $i$-th word representation.

\subsubsection{Joint-task Multi-Label Supervised Contrastive Learning}
\begin{figure}[t]
 \centering
 \includegraphics[width = 0.48\textwidth]{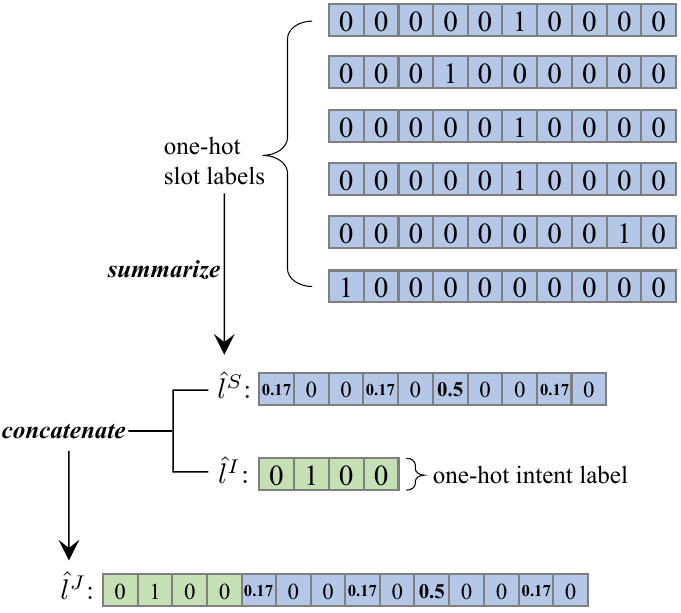}
 \caption{Illustration of the process of constructing the sentence-level joint-task label using the one-hot intent labels and the one-hot slot labels (non-O). 
 We first summarize the one-hot slot labels into a single sentence-level soft slot label, which is then concatenated with the one-hot intent label to form the final sentence-level joint task label.
 In this example, w.l.o.g, there are total 4 different intents and 10 different slots.The reason we do not count the O slot is that it does not convey any semantics information.}
 \label{fig: slotlabel}
\end{figure}

To jointly model intent detection and slot filling in supervised contrastive learning, we have to construct the sentence-level joint-task label by ourselves, because it is not provided in the dataset.
To this end, we first obtain the sentence-level slot label $\hat{l}^S$ by summarizing all non-O slot labels in one label vector:
\begin{equation}
\hat{l}^{S} = \frac{\sum^N_{i = 1, l^s_i \neq \text{O}}  \hat{l}^s_i}{\sum^N_{i = 1, l^s_i \neq \text{O}}{1}}
\end{equation}
where ${l}^s_i$ denotes the slot label if $i$-th word, and $\hat{l}^s_i$ denotes the one-hot label.
$\hat{l}^S$ can represent the slot-specific semantics of the utterance.
Then we concatenate $\hat{l}^S$ and the one-hot intent label $\hat{l}^I$ to form the joint-task label $\hat{l}^{J}$.
An example of obtaining $\hat{l}^{J}$ is shown in Fig. \ref{fig: slotlabel}. 

Since $\hat{l}^{J}$ is not a one-hot label, this contrastive learning mechanism performs multi-label supervised contrastive learning.
In this case, some samples may share some common labels with the current utterance, while others may share all labels or no one.
Therefore, how to measure the golden similarity of the two instances is the key challenge.
In this paper, we propose to use the Hadamard product with normalization to achieve this.
Denoting the current utterance as $i$, the whole process of this contrastive learning mechanism can be formalized as follows:
\begin{equation}
\begin{split}
\mathcal{L}_{\text{slscl}}^{\text{Joint}} &= -\sum^K_k{w_{ik}\operatorname{log}\frac{e^{s(h^i_{\text{CLS}}, h^k_{\text{CLS}})}}{\sum^{K}_j{e^{s(h^i_{\text{CLS}}, h^j_{\text{CLS}})}}}}\\
w_{ik} &= \frac{\mu_{ik}}{\sum^K_j{\mu_{ij}}}\\
\mu_{ij} &= \hat{l}^{J}_{i} \odot \hat{l}^{J}_{j}
\end{split}
\end{equation}
A large $\mu_{ik}$ denotes the sample $k$ is quite similar to the anchor and leads to a large $w_{ik}$ assigned to the loss function to pull them closer.
Instead, $\mu_{ik}=0$ denotes they have totally different labels.
Since their distance only appears in the denominator, they will be pushed apart by the negative gradient.

\subsection{Cross-lingual Supervised Contrastive Learning}
Generally, the model can learn relatively high-quality source language representations from the training set.
To transfer the label-aware semantics knowledge from the source language to the target languages, we propose the cross-lingual supervised contrastive learning mechanism to perform explicit label-aware cross-lingual semantics alignment.
The current utterance's multilingual view is regarded as the anchor, and its sentence (word) representations are drawn together with the source language positive samples in $Q^u$ ($Q^w$) that share the same/similar label(s), while they are distinguished from the negative samples that have different labels.
\subsubsection{Intent Supervised Contrastive Learning}
Denoting current utterance as $i$, similar to Eq. 4, this contrastive learning mechanism can be formulated as follows:
\begin{equation}
\begin{split}
\mathcal{L}_{\text{clscl}}^{\text{I}} &= -\sum^{K}_k \frac{\mu_{ik}}{\sum^K_j {\mu_{ij}}}\operatorname{log}\frac{{ e^{s(h^{[\text{ml},i]}_{\text{CLS}}, h^k_{\text{CLS}})}}}{\sum^{K}_k{e^{s(h^{[\text{ml},i]}_{\text{CLS}}, h^{k}_{\text{CLS}})}}}\\
\mu_{ik} &= \hat{l^I_i}\odot \hat{l^I_k}
\end{split}
\end{equation} 
where  $\mu_{ik}$ equals 0 or 1, indicating whether the $k$-th sample in $Q^u$ is a positive sample of the current utterance's sentence representation.
$h^{[\text{ml},i]}_{\text{CLS}}$ denotes the sentence representation of the current utterance's multilingual view. 

\subsubsection{Slot Supervised Contrastive Learning}
Similar to Eq. 5, this contrastive learning mechanism can be formulated as follows:
\begin{equation}
 \begin{split}
\mathcal{L}_{\text{clscl}}^{\text{S}} &= -\frac{1}{n^2}\sum^n_{i}\sum^n_{j} \sum^{K}_k  \frac{\mu_{i}^{[k,j]}}{\sum_a^K\mu_{ia}} \operatorname{log}\frac{ {e^{s(h^{\text{ml}}_i, h_j^k )}}}{\sum^{K}_a {e^{s(h^{\text{ml}}_i, h^a_j)}}}\\
\mu_{i}^{[k,j]} &= \hat{l^s_i}\odot \hat{l}^s_{[k,j]}
\end{split}
\end{equation}
where $\mu_{i}^{[k,j]}$ equals 0 or 1, indicating whether the $j$-th word representation of the $k$-th sample in $Q^w$ is a positive sample of the current utterance's $i$-th word representation.
$h^{\text{ml}}_i$ denotes the $i$-th word representation of the current utterance's multilingual view. 

\subsubsection{Joint-task Multi-Label Supervised Contrastive Learning}
Similar to Eq. 6, this contrastive learning mechanism can be formulated as:
\begin{equation}
\begin{split}
\mathcal{L}_{\text{clscl}}^{\text{Joint}} &= -\sum^K_k{w_{ik}\operatorname{log}\frac{e^{s(h^{[\text{ml},i]}_{\text{CLS}}, h^k_{\text{CLS}})}}{\sum^{K}_j{e^{s(h^{[\text{ml},i]}_{\text{CLS}},  h^j_{\text{CLS}})}}}}\\
 w_{ik} &= \frac{\mu_{ik}}{\sum^K_j{\mu_{ij}}}\\
 \mu_{ij} &= \hat{l}^{J}_{i} \odot \hat{l}^{J}_{j}
\end{split}
\end{equation}

\subsection{Multilingual Supervised Contrastive Learning}
Although we cannot directly perform supervised contrastive learning on the target languages due to the lack of training data, we propose the multilingual supervised contrastive learning to achieve the pseudo target language supervised contrastive learning inspired by the two facts: (1) we have the representations of the multilingual view, which contains the semantics of target languages;  (2) multilingual views also have ground-truth labels because each one shares the same labels with its original source language utterance.
This contrastive learning mechanism aims to pull the current multilingual view's sentence (word) representation together with the samples in $Q^u_{ml}$ ($Q^w_{ml}$) or push them apart regarding whether they share the same/similar label(s).
\subsubsection{Intent Supervised Contrastive Learning}
Denoting the current source language utterance as $i$, this contrastive learning mechanism can be formulated as follows:
\begin{equation}
\begin{split}
\mathcal{L}_{\text{mlscl}}^{\text{I}} &= -\sum^{K}_k \frac{\mu_{ik}}{\sum^K_j {\mu_{ij}}}\operatorname{log}\frac{{e^{s(h^{[\text{ml},i]}_{\text{CLS}}, h^{[\text{ml},k]}_{\text{CLS}})}}}{\sum^{K}_j{e^{s(h^{[\text{ml},i]}_{\text{CLS}}, h^{[\text{ml},j]}_{\text{CLS}})}}} \\
\mu_{ik} &= \hat{l}^I_i \odot \hat{l}^I_k
\end{split}
\end{equation}
 where $\mu_{ik}$ equals 0 or 1, indicating whether the $k$-th sample in $Q^u_{ml}$ is a positive sample of the current utterance's multilingual view's sentence representation.

\subsubsection{Slot Supervised Contrastive Learning}
Similar to Eq. 5, this contrastive learning mechanism can be formulated as follows:
\begin{equation}
\begin{split}
\mathcal{L}_{\text{mlscl}}^{\text{S}} &= -\frac{1}{n^2}\sum^n_{i}\sum^n_{j}\sum^{K}_k  \frac{\mu_{i}^{[k,j]}}{\sum_a^K \mu_{ia}} \operatorname{log}\frac{{ e^{s(h_i^{\text{ml}}, h^{[\text{ml},k]}_j)}}}{\sum^{K}_a{e^{s(h_i^{\text{ml}}, h^{[\text{ml},a]}_j)}}}\\
\mu_{i}^{[k,j]} &= \hat{l^s_i}\odot \hat{l^s_{[k,j]}}
\end{split}
\end{equation}
where $\mu_{i}^{[k,j]}$ equals 0 or 1, indicating whether the $j$-th word representation of the $k$-th sample in $Q^w_{ml}$ is a positive sample of the current utterance's multilingual view's $i$-th word representation. 

\subsubsection{Joint-task Multi-Label Supervised Contrastive Learning}
Denoting the current source language utterance as $i$, this contrastive learning mechanism can be formulated as follows:
\begin{equation}
\begin{split}
\mathcal{L}_{\text{mlscl}}^{\text{Joint}} &= -\sum^K_k{w_{ik}\operatorname{log}\frac{e^{s(h^{[\text{ml}, i]}_{\text{CLS}}, h^{[\text{ml},k]}_{\text{CLS}})}}{\sum^{K}_j{e^{s(h^{[\text{ml}, i]}_{\text{CLS}}, h^{[\text{ml},j]}_{\text{CLS}})}}}}\\
w_{ik} &= \frac{\mu_{ik}}{\sum^K_j{\mu_{ij}}}\\
\mu_{ij} &= \hat{l}^{J}_{i} \odot \hat{l}^{J}_{j}
\end{split}
\end{equation}

\subsection{Training Objective}
Denoting $\mathcal{L}_I$ and $\mathcal{L}_S$ as the standard loss function for intent detection and slot filling, the final training objective of HC$^2$L is the weighted sum of $\mathcal{L}_I$, $\mathcal{S}$ and all the above contrastive learning objectives:
\begin{equation}
\begin{split}
&\mathcal{L} = \lambda_I \mathcal{L}_I + \lambda_S \mathcal{L}_S + \mathcal{L}_{\text{un}} + \mathcal{L}_{\text{slscl}} + \mathcal{L}_{\text{clscl}} +  \mathcal{L}_{\text{mlscl}} \\
&\mathcal{L}_{\text{un}} = \lambda^I_{\text{un}} \mathcal{L}^{I}_{un} + \lambda^S_{\text{un}} \mathcal{L}^{\text{S}}_{\text{un}}+  \lambda^{\text{GIS}}_{\text{un}} \mathcal{L}^{\text{GIS}}_{\text{un}} \\
&\mathcal{L}_{\text{slscl}} = \beta_I \mathcal{L}_{\text{slscl}}^{\text{I}} + \beta_S \mathcal{L}_{\text{slscl}}^{\text{S}} + \beta_{J}\mathcal{L}_{\text{slscl}}^{\text{Joint}} \\
&\mathcal{L}_{\text{clscl}} = \gamma_1  (\beta_I \mathcal{L}_{\text{clscl}}^{\text{I}} + \beta_S \mathcal{L}_{\text{clscl}}^{\text{S}} + \beta_{J}\mathcal{L}_{\text{clscl}}^{\text{Joint}})\\
&\mathcal{L}_{\text{mlscl}} = \gamma_2  (\beta_I \mathcal{L}_{\text{mlscl}}^{\text{I}} + \beta_S \mathcal{L}_{\text{mlscl}}^{\text{S}} + \beta_{J}\mathcal{L}_{\text{mlscl}}^{\text{Joint}})
\end{split}
\end{equation}
where $\lambda_*$, $\beta_*$ and $\gamma_*$ are hyper-parameters balancing the loss terms.
The standard loss functions for intent detection ($\mathcal{L}_I$) and slot filling ($\mathcal{L}_S$) in Eq.13 are defined as following:
\begin{equation}
\begin{split}
\mathcal{L}_I & =-\sum_{j=1}^{C_I} \hat{\mathbf{y}}^I[j] \log \left(L^I[j]\right) \\
\mathcal{L}_S & =-\sum_{i=1}^n \sum_{j=1}^{C_S} \hat{\mathbf{y}}^s_i[j] \log(l_i^s[j])
\end{split}
\end{equation}
where $C_I$ and $C_S$ are the sets of intent labels and slot labels, respectively;
$\hat{\mathbf{y}}^I$ and $\hat{\mathbf{y}}^s$ are the ground-truth intent labels and slot labels, respectively.

\section{Experiments}
\subsection{Settings}
\subsubsection{Dataset}
Following previous works, we evaluate our model on the multilingual benchmark dataset of MultiATIS++ \cite{slot-align-and-rec}.
This dataset includes 9 languages: English (en), Spanish (es), Portuguese (pt), German (de), French (fr), Chinese (zh), Japanese (ja), Hindi (hi) and Turkish (tr).
The dataset includes 18 kinds of intents and 84 kinds of slot labels.
The training set is in English, consisting of 4488 samples.
The validation set for each language includes 490 samples, except for Hindi and Turkish, which have 160 and 60 validation samples, respectively.
The testing set for each language includes 893 samples, except for Turkish, which has 715 testing samples.

\begin{table*}[t]
\centering
\fontsize{8}{10}\selectfont
\caption{Main results. Our model significantly outperforms baselines with $p<0.05$ under the t-test.} 
\setlength{\tabcolsep}{3mm}{
\begin{tabular}{l|ccccccccc|c}
\hline
\textbf{Intent Accuracy} & en & de & es & fr & hi & ja & pt & tr & zh & Avg.   \\ \hline
mBERT \cite{slot-align-and-rec} & - & 95.27 & 96.35 & 95.92 & 90.96 & 79.42 & 94.96 & 69.59 & 86.27  &- \\
mBERT \cite{bert} &98.54 &95.40 &96.30& 94.31 &82.41 &76.18 &94.95 &75.10 &82.53 & 88.42 \\
Ensemble-Net \cite{crossing-the-conversational-chasm}&90.26  &92.50 &96.64 &95.18 &77.88 &77.04 &95.30& 75.04 &84.99 & 87.20 \\
CoSDA \cite{cosda-ml}& 95.74 &94.06 &92.29& 77.04 &82.75 &73.25& 93.05& 80.42& 78.95 & 87.32 \\
GL-CLEF \cite{gl-clef} & 98.54& 98.09 &97.91 &97.72 &86.34 &80.02 &96.41& 81.82 &88.24 & 91.68 \\ \hline
HC$^2$L (ours) &\textbf{99.10} &\textbf{98.88} &\textbf{99.02} &\textbf{99.12} & \textbf{90.37} &\textbf{88.83} &\textbf{97.87} &\textbf{89.23} &\textbf{93.06} & \textbf{95.05} \\
\hline
\hline
\textbf{Slot F1} & en & de & es & fr & hi & ja & pt & tr & zh & Avg.   \\ \hline
mBERT \cite{slot-align-and-rec} &- &82.61 &74.98 &75.71 &31.21& 35.75 &74.05 &23.75& 62.27&- \\
mBERT \cite{bert} &95.11& 80.11 &78.22 &82.25 &26.71 &25.40 &72.37 &41.49 &53.22& 61.66  \\
Ensemble-Net \cite{crossing-the-conversational-chasm}&85.05 & 82.75 &77.56 &76.19 &14.14 &9.44 &74.00 &45.63& 37.29&55.78 \\
CoSDA \cite{cosda-ml}& 92.29& 81.37 &76.94 &79.36& 64.06 &66.62 &75.05 &48.77 &77.32 & 73.47 \\
GL-CLEF \cite{gl-clef} & 95.89 &84.29 &85.76 &85.85 &65.55 &66.36 &81.50 &68.34& 78.30 & 79.09 \\ \hline
HC$^2$L (ours) &\textbf{96.18} &\textbf{90.17} & \textbf{87.86}  & \textbf{88.03}  & \textbf{71.91}  & \textbf{75.78} & \textbf{83.49}  & \textbf{71.29} & \textbf{81.96} & \textbf{82.96} \\ 
\hline
\hline
\textbf{Overall Accuracy} & en & de & es & fr & hi & ja & pt & tr & zh & Avg.   \\ \hline
mBERT \cite{bert} & 87.12 & 52.69 & 52.02 & 37.29 & 4.92 & 7.11 & 43.49 & 4.33 & 18.58 & 36.29 \\
AR-S2S-PTR \cite{ars2sptr} &86.83 &34.00 &40.72 &17.22 &7.45 &10.04 &33.38 &– &23.74& - \\
IT-S2S-PTR \cite{its2sptr} & 87.23 &39.46 &50.06& 46.78 &11.42& 12.60 &39.30 &– &28.72& - \\
CoSDA \cite{cosda-ml}&77.04 &57.06& 46.62 &50.06& 26.20 &28.89 &48.77 &15.24 &46.36 & 44.03\\
GL-CLEF \cite{gl-clef} &88.69& 66.26& 63.71 & 60.05&26.76 &32.84 &60.54 &30.35& 53.08& 53.56 \\ \hline
HC$^2$L (ours) &\textbf{89.92} &\textbf{72.20} &\textbf{66.05} &\textbf{67.51} &\textbf{34.83} &\textbf{42.44} &\textbf{63.34}&\textbf{36.50} &\textbf{58.01} & \textbf{58.98} \\ 
\hline
\end{tabular}}
\label{table: results}
\end{table*}

\subsubsection{Implementation}
Following previous models, we adopt the base case mBERT as the encoder.
(e.g., learning rate, batch size, dropout rate, sample queue size, $\lambda_I$, $\lambda_S$, $\lambda^I_{\text{un}}$,
$\lambda^S_{\text{un}}$ and $\lambda^{\text{GIS}}_{\text{un}}$).
We set the learning rate as 5e-6. The batch size is 32.
Dropout rate is 0.1.
 $\lambda_I$ and $\lambda_S$ are set as 1.
  $\lambda^I_{\text{un}}$, $\lambda^S_{\text{un}}$ and $\lambda^{\text{GIS}}_{\text{un}}$ are set as 0.01, 0.005 and 0.01, respectively.
For fair comparisons, all the above hyper-parameters are set as the same as GL-CLEF \cite{gl-clef}. 
In our experiments, we only tune $\beta_I$, $\beta_S$, $\beta_J$, $\gamma_1$ and $\gamma_2$, which balance the loss terms corresponding to the three kinds of supervised CL proposed in this paper.
The hyper-parameters tested in training our models are listed in Table \ref{table: hyper-parameter}. We test all combinations of them and choose the one achieving the highest average of all the 9 languages' overall accuracies.

\begin{table}[h]
\centering
\fontsize{8}{10}\selectfont
\caption{Tuned hyper-parameters. Finally, chosen values are in \textbf{bold}.} 
\setlength{\tabcolsep}{4mm}{
\begin{tabular}{c|c}
\toprule
\textbf{Hyper-parameters} & \textbf{Values} \\ \midrule
$\beta_{I}$ &  1e-5, 1e-4, 1e-3, \textbf{1e-2}, 0.1 \\
$\beta_S$ & 1e-5, \textbf{1e-4}, 1e-3, 1e-2, 0.1  \\
$\beta_J$ & 1e-5, \textbf{1e-4}, 1e-3, 1e-2, 0.1 \\
$\gamma_1$ & 1e-2, \textbf{0.1}, 1.0 \\
$\gamma_2$ & 1e-2, \textbf{0.1}, 1.0   \\ \bottomrule
\end{tabular}}
\label{table: hyper-parameter}
\end{table}
We select the best-performing model on the validation set and report its performance on the test set.
All experiments in this work are conducted on a single NVIDIA A100 80G GPU.

\subsubsection{Baseline}
We compare our model with the following baselines: \\
(1) mBERT \cite{bert}. It is based on the same model architecture as BERT \cite{bert} and adopts the same training procedure. Its training data covers the Wikipedia pages of 104 languages with a shared subword vocabulary. Therefore, mBERT can obtain the share embeddings across languages, which facilitate cross-lingual NLP tasks. \\
(2) Ensemble-Net \cite{crossing-the-conversational-chasm}. The final predictions of this model are determined by 8 independent models through majority voting. Each model is separately trained on a single source language. \\
(3) CoSDA \cite{cosda-ml}. This model proposes a novel data augmentation framework to generate multi-lingual code-switching data that is used to to fine-tune mBERT. CoSDA can align representations from source and multiple target languages.\\
(4) GL-CLEF \cite{gl-clef}. This model proposes to leverage the unsupervised contrastive learning to perform explicit semantics alignment between the utterance and its multilingual view obtained by code-switching. \\

For fair comparisons, we reproduce GL-CLEF's results with the default hyper-parameters.

\subsection{Main Results}

Following previous works, we adopt accuracy, F1 score and overall accuracy for the metrics to evaluate intent detection, slot filling and sentence-level semantics frame parsing, respectively.

The main results comparison based on mBERT encoder is shown in Table \ref{table: results}.
We can observe that our HC$^2$L model significantly outperforms all baselines by a large margin.
In terms of overall accuracy, our model obtains a relative improvement of 10.1\% over the up-to-date best model GL-CLEF.
This demonstrates that our proposed three supervised contrastive learning mechanisms can comprehensively improve the semantics via performing explicit label-aware alignments for the source language, cross-lingual and multilingual scenarios.
Besides, we can find that the improvements on languages having inferior performances are sharper.
For instance, HC$^2$L achieves nearly 50\% relative improvement in terms of overall accuracy for Hindi (hi).
This proves that our proposed cross-lingual supervised contrastive learning can effectively transfer the label-aware semantics knowledge from the source language  into target languages, and the multilingual supervised contrastive learning can achieve pseudo target language supervised contrastive learning, which can learn better and more discriminative target language semantics.
%

\subsection{Ablation Study}
We conduct extensive ablation experiments to study the effect of the contrastive learning mechanisms in our HC$^2$L model.
\subsubsection{Effect of Unsupervised Contrastive Learning and Supervised Contrastive Learning}
\begin{figure}[t]
 \centering
 \includegraphics[width = 0.46\textwidth]{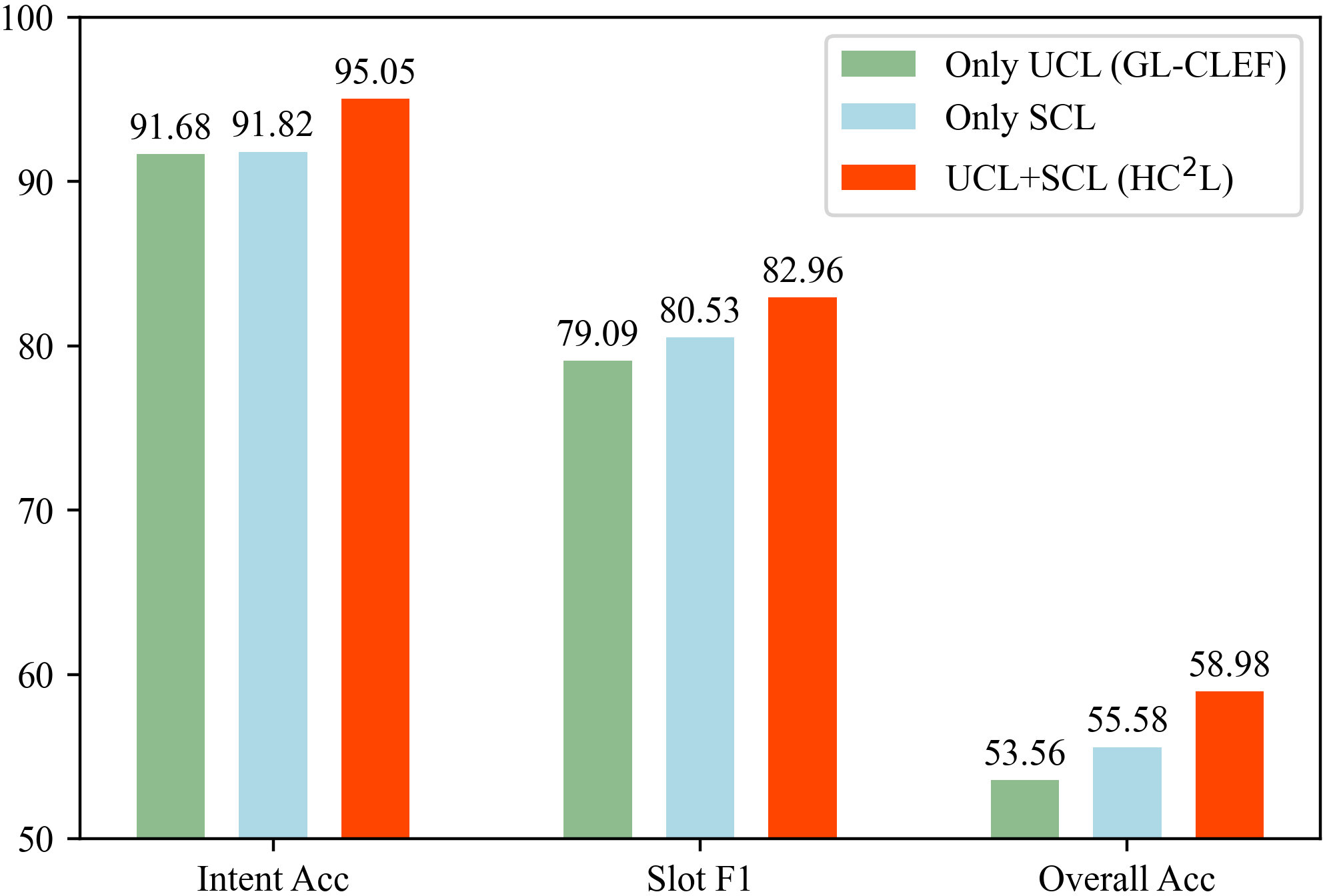}
 \caption{Ablation results on Only UCL, Only SCL and UCL+SCL. Only UCL is a variant that only performs unsupervised contrastive learning, which is equal to the GL-CLEF model. Only SCL is a variant that only performs the source language, cross-lingual and multilingual supervised contrastive learning. UCL+SCL denotes all unsupervised and supervised contrastive learning mechanism are adopted, which is equal to our HC$^2$L model.}
 \label{fig: ablat_clandscl}
\end{figure}

Fig. \ref{fig: ablat_clandscl} show the results of the ablation experiments for studying the effect of unsupervised contrastive learning and supervised contrastive learning.
Firstly, we can observe that Only SCL outperforms Only UCL on all metrics, proving that our proposed supervised contrastive learning mechanisms contribute more than the cross-lingual unsupervised contrastive learning.
This can be attributed to the fact that our proposed three kinds of supervised CL can comprehensively capture the label-aware semantic structure and enhance the label-aware knowledge transfer.
Besides, UCL+SCL, namely our HC$^2$L model, achieves the best performance.
This proves that the unsupervised contrastive learning and our proposed three kinds of supervised contrastive learning can effectively cooperate with each other to learn better semantic representations in the training procedure.
\subsubsection{Effect of Different Kinds of Supervised Contrastive Learning}
\begin{figure}[t]
 \centering
 \includegraphics[width = 0.46\textwidth]{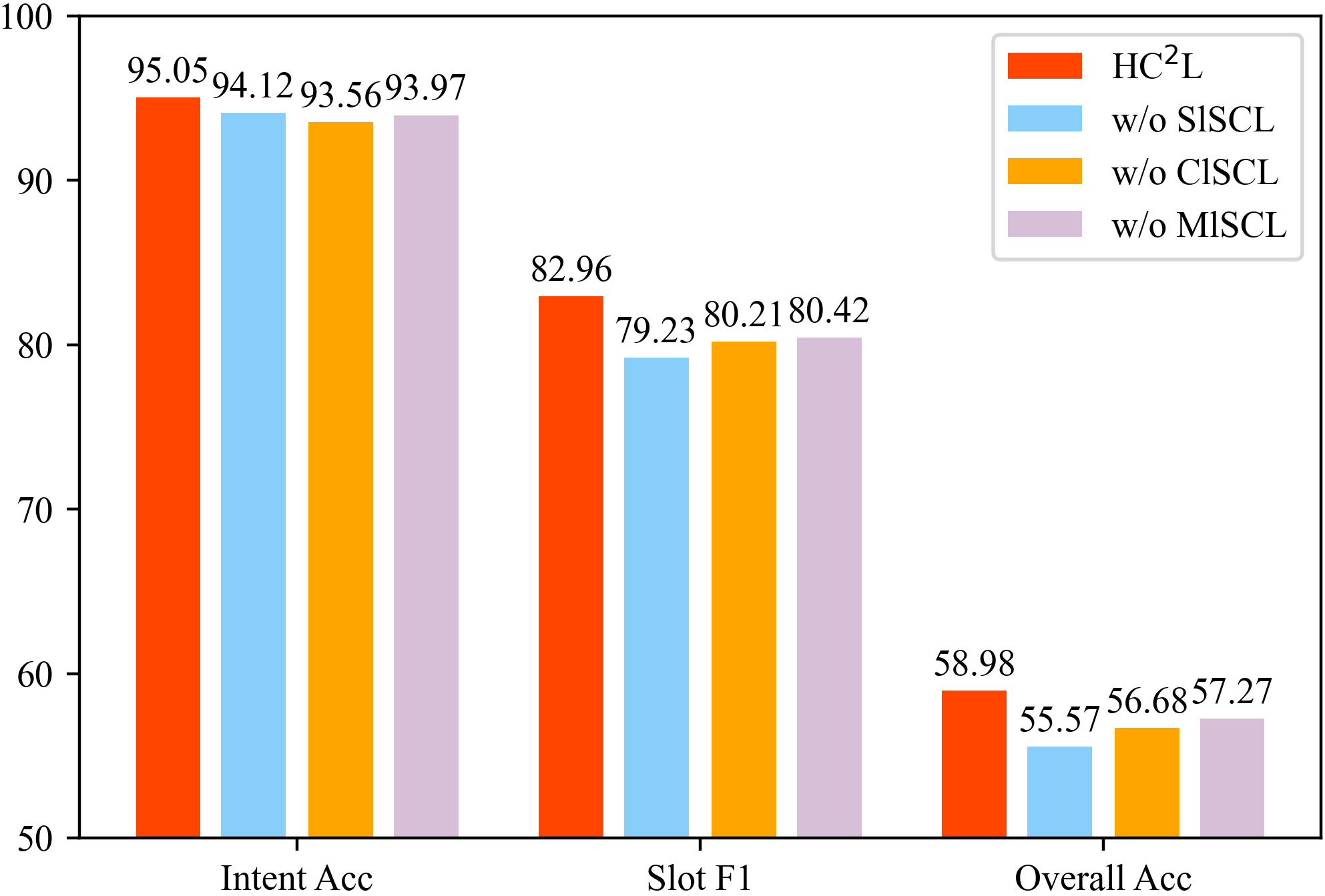}
 \caption{Ablation results on the three kinds of supervised contrastive learning. SlSCL denotes source language supervised contrastive learning, ClSCL denotes cross-lingual supervised SCL, and MlSCL denotes multilingual supervised contrastive learning.}
 \label{fig: ablat_scl}
\end{figure}
We conduct a group of ablation experiments to study the effectiveness of each kind of supervised CL proposed in this paper.
The results are shown in Fig. \ref{fig: ablat_scl}.
We can find that all three kinds of supervised contrastive learning bring significant improvements.
This is because source language supervised CL can perform explicit label-aware semantics alignments for the source language.
And cross-lingual supervised contrastive learning can sufficiently transfer the label-aware semantics knowledge from the source language into target languages.
As for multilingual supervised contrastive learning, it can further learn better and more discriminative multilingual representations via drawing together and pushing part the semantics of target languages regarding whether they share the same/similar labels.
\begin{figure}[t]
 \centering
 \includegraphics[width = 0.46\textwidth]{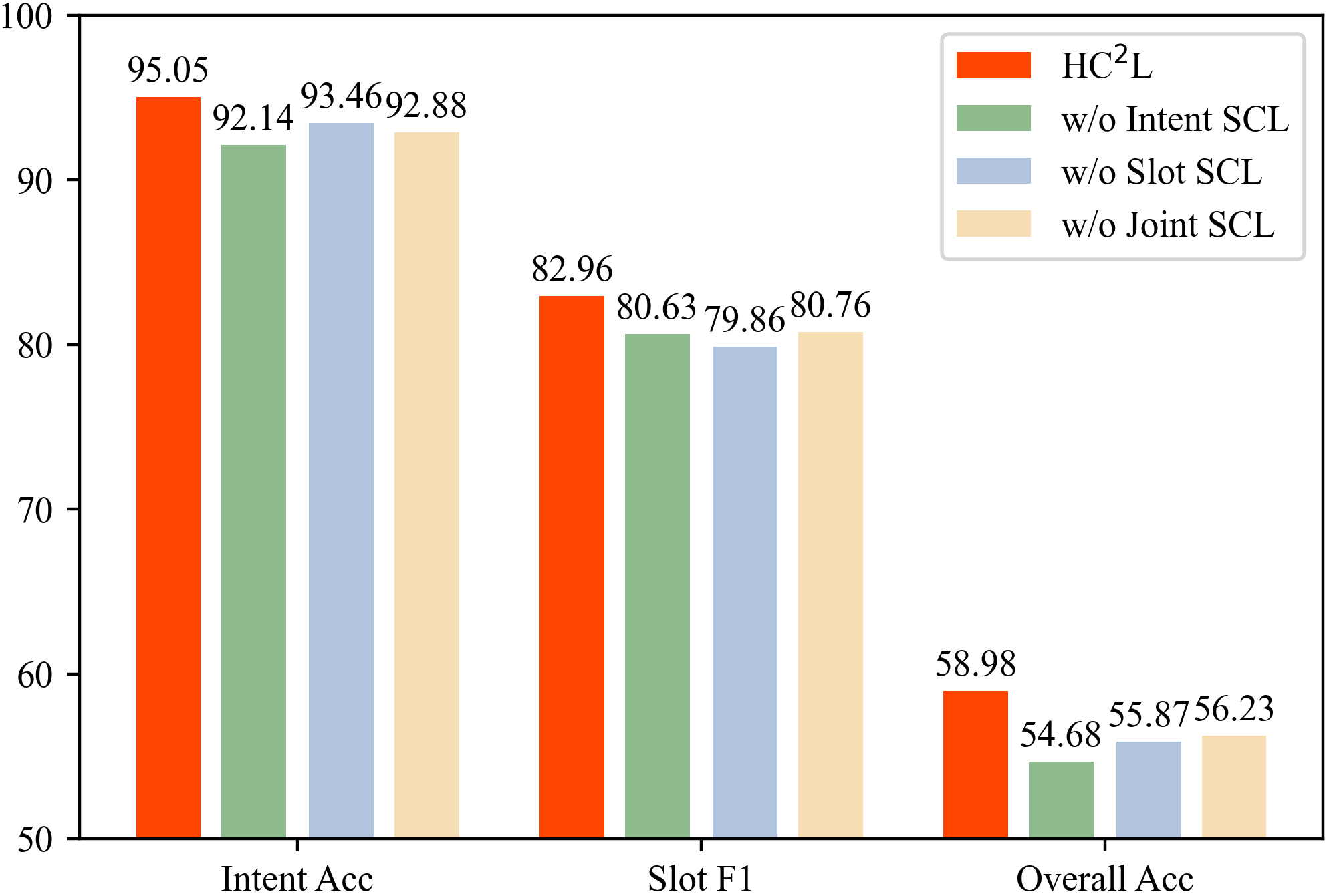}
 \caption{Ablation results on the single-task supervised contrastive learning and joint-task supervised contrastive learning. }
 \label{fig: ablat_cl}
\end{figure}

\begin{figure*}[t]
 \centering
 \includegraphics[width = \textwidth]{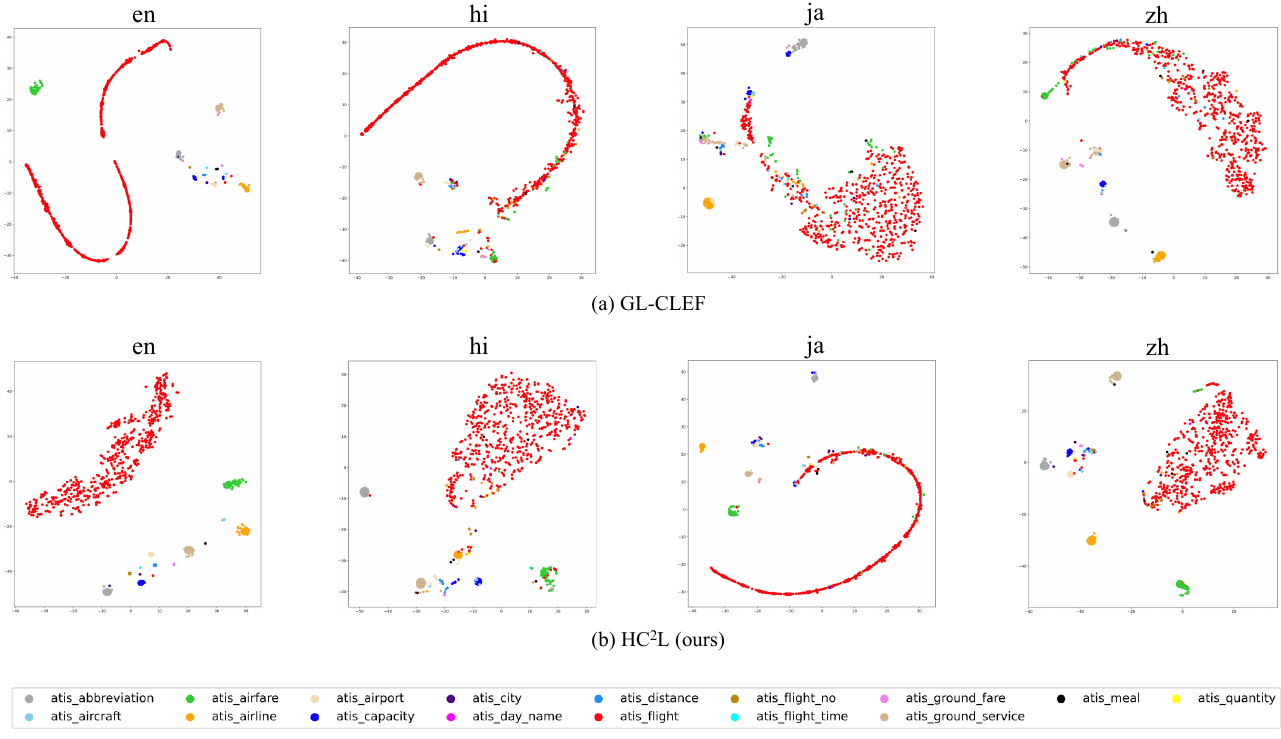}
 \caption{TSNE visualization of the sentence embeddings of the source language (en) and three target languages (hi, ja, zh). Different colors denote the sentence embeddings correspond to different intent classes.}
 \label{fig: classcluster}
\end{figure*}

\begin{table*}[t]
\centering
\fontsize{8}{10}\selectfont
\caption{Results based on XLM-R. Our model significantly outperforms baselines with $p<0.05$ under the t-test.} 
\setlength{\tabcolsep}{3mm}{
\begin{tabular}{l|ccccccccc|c}
\hline
\textbf{Intent Accuracy} & en & de & es & fr & hi & ja & pt & tr & zh & Avg.   \\ \hline
XLM-R \cite{xlm-r}  &98.32 &97.19 &98.03 &94.94 &88.91 &88.50 &96.41 &72.45 &91.15 & 93.02\\
XLM-R + GL-CLEF \cite{gl-clef}   &98.77 & 97.87 &98.28 & 97.72 & 85.33 & 83.52 & 97.65 & 85.03 &90.59 & 92.75\\ \hline
XLM-R + HC$^2$L (ours) & \textbf{99.22}  & \textbf{98.43}  & \textbf{97.42}  & \textbf{97.72}  & \textbf{92.61} & \textbf{91.20}  & \textbf{97.98} & \textbf{86.15} & \textbf{92.95} &\textbf{94.85} \\
\hline
\hline
\textbf{Slot F1} & en & de & es & fr & hi & ja & pt & tr & zh & Avg.   \\ \hline
XLM-R \cite{xlm-r}&94.58 &72.35 &76.72 &71.81 &60.51 &9.31 &70.08 &45.21 &13.44 & 57.38 \\
XLM-R + GL-CLEF \cite{gl-clef}   &95.81 & 87.52 &87.51 &82.67 & 67.64 & 65.15 & 78.66 & 55.34 & 80.69 & 75.73 \\
XLM-R + HC$^2$L (ours)  &\textbf{95.83} & \textbf{87.32} & \textbf{87.76} & \textbf{83.30} & \textbf{73.65}  & \textbf{72.84} & \textbf{79.24} & \textbf{57.12} & \textbf{81.32} & \textbf{77.74} \\
\hline
\hline
\textbf{Overall Accuracy} & en & de & es & fr & hi & ja & pt & tr & zh & Avg.   \\ \hline
XLM-R \cite{xlm-r}  &87.45 &43.05 &42.93 &43.74 &19.42 &5.76 &40.80 &9.65 &6.60 & 33.31 \\
XLM-R + GL-CLEF \cite{gl-clef}           &89.03 &68.16 &63.96 &60.68 & 30.80 & 28.10 & 55.94 &18.74 & 58.79 & 52.69   \\ \hline
XLM-R + HC$^2$L (ours) &\textbf{89.36} & \textbf{68.16}  & \textbf{64.70}  & \textbf{61.57}  & \textbf{39.20}  & \textbf{44.13}  & \textbf{57.51}  & \textbf{20.98}  & \textbf{58.90} & \textbf{56.06}  \\ 
\hline
\end{tabular}}
\label{table: xlr-results}
\end{table*}

\subsubsection{Effect of Single-task and Joint-task Supervised Contrastive Learning}

We also conduct ablation experiments to study the single-task supervised contrastive learning (Intent SCL and Slot SCL) and the joint-task supervised contrastive learning (Joint SCL).
The results are shown in Fig. \ref{fig: ablat_cl}.
Intent SCL and Slot SCL use the provided ground-truth labels as the supervision signal to perform supervised contrastive learning.
And we can observe that both of them have significant contributions.
As for joint-task supervised contrastive learning, since there is no provided label, we construct the sentence-level joint-task labels by ourselves and use them to perform multi-label supervised contrastive learning.
The performance gap between the full model (HC$^2$L) and \textit{w/o Joint SCL} demonstrates the effectiveness of Joint SCL, which can capture the dual-task correlations and jointly model the two tasks in supervised contrastive learning for zero-shot cross-lingual SLU.

\subsection{Effect of Multilingual Encoder}

To verify our method can still work effectively based on other pre-trained multilingual encoders, we conduct experiments based on XLM-R \cite{xlm-r} using the same hyper-parameters of mBERT+HC$^2$L.
The results are shown in Table \ref{table: xlr-results}. 
We can find that our model can bring a large improvement of 68.3\% to XLM-R, and our model significantly outperforms GL-CLEF by 6.4\%.
This proves that the contributions of our proposed supervised contrastive learning mechanisms are model-agnostic, it can always bring improvements to different multilingual encoders.
In the future, if a stronger pre-trained multilingual encoder is proposed, our model can still be applied to further improve the performance.

\subsection{Visualization of Sentence Embeddings}

\subsubsection{Intent Clustering}

We also visualize the sentence embeddings of test samples in different intent classes, as shown in Fig. \ref{fig: classcluster}.
Compared with GL-CLEF, our HC$^2$L can generate intent clusters exhibiting clearer separation.
Although GL-CLEF can disentangle different classes for \texttt{en}, our models’ intent clusters are clearer and farther away from each other than the ones of GL-CLEF. This can prove that our model can better capture the label-aware semantics structure in the source language, which can be attributed to our proposed source language supervised CL. As for the target languages, our model makes different intent clusters better separated apart than GL-CLEF. Taking \texttt{ja} as an example, we can find that compared to the generated sentence embeddings of our model, the ones of GL-CLEF can hardly form intent clusters. Especially, the sentence embeddings of GL-CLEF corresponding to intent ‘atis\_airfare’ and ‘atis\_flight’ obviously overlap with each other, while the sentence embeddings of our model corresponding to intent ‘atis\_airfare’ form a clear cluster and far away from other intent clusters. The above observations prove that our model can generate more discriminative semantics representations for different labels, thanks to our proposed novel supervised CL mechanisms.

\begin{figure*}[t]
 \centering
 \includegraphics[width = 0.8\textwidth]{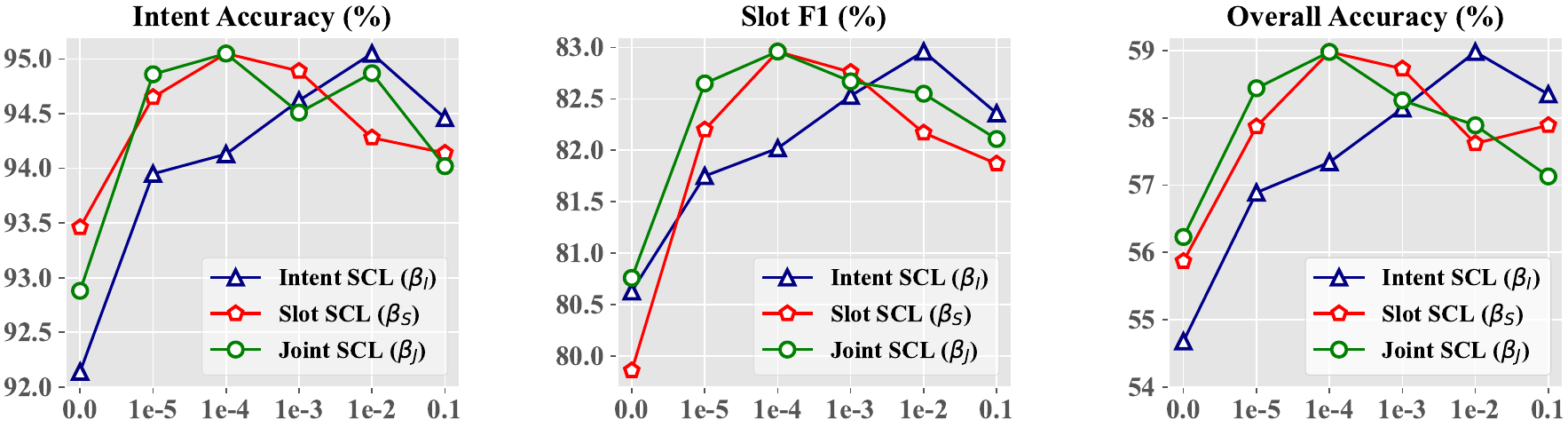}
 \caption{The performances of our HC$^2$L model on different values of $\beta_I$, $\beta_S$, $\beta_J$.}
 \label{fig: lambda}
\end{figure*}

\begin{figure*}[t]
 \centering
 \includegraphics[width = 0.8\textwidth]{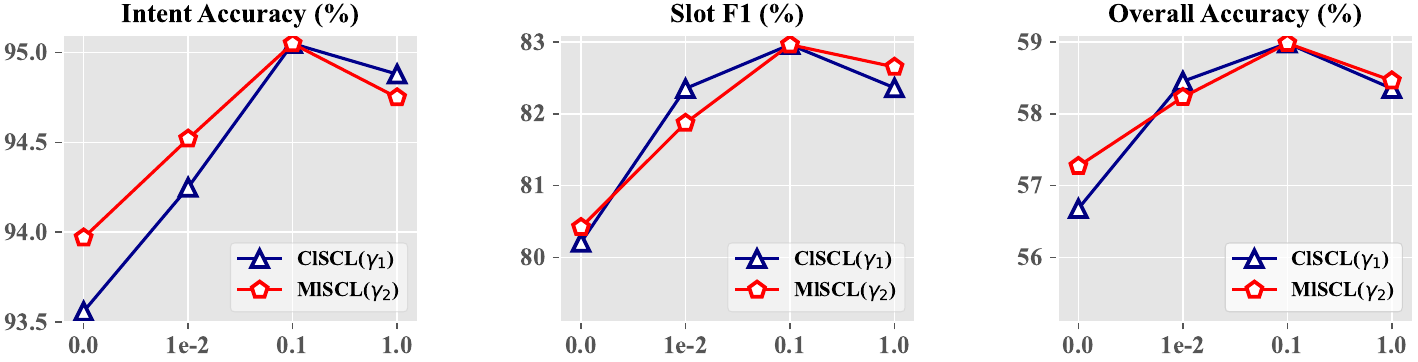}
 \caption{The performances of our HC$^2$L model on different values of $\gamma_1$ and $\gamma_2$. ClSCL denotes cross-lingual supervised contrastive learning and MlSCL denotes multilingual supervised contrastive learning.}
 \label{fig: beta}
\end{figure*}

\begin{figure}[t]
 \centering
 \includegraphics[width = 0.48\textwidth]{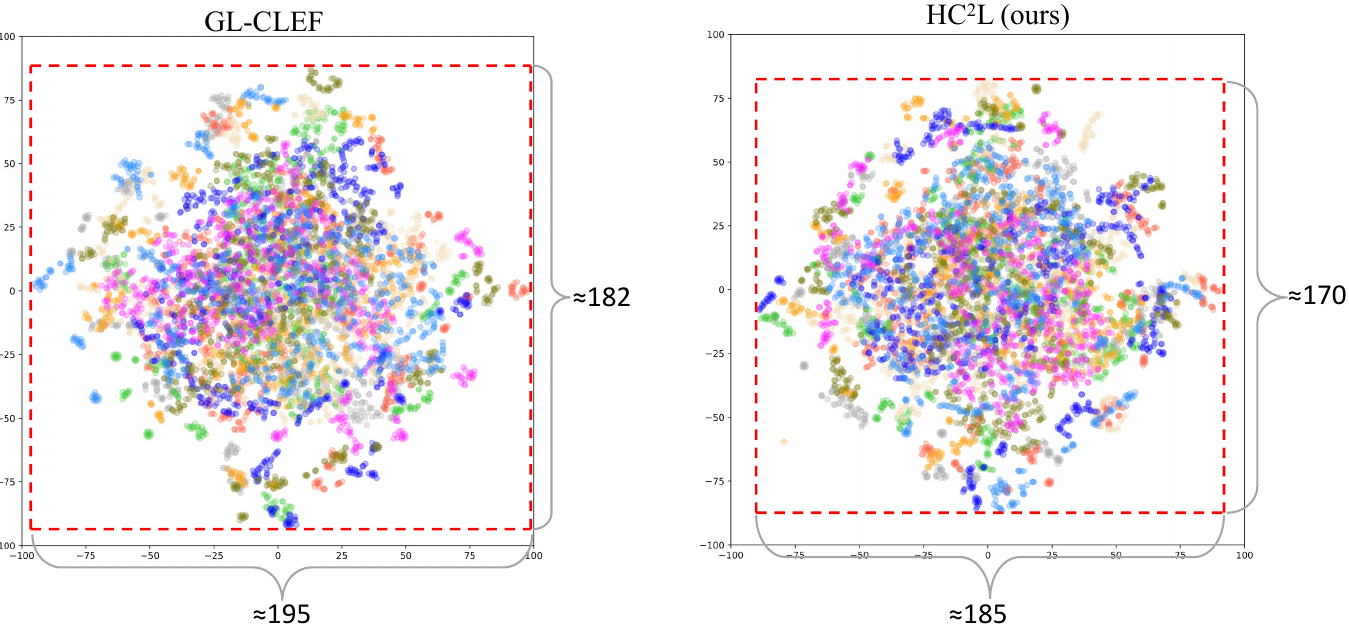}
 \caption{TSNE visualization of sentence embeddings generated by GL-CLEF and our HC$^2$L model. Different colors denote different languages. All embeddings are normalized on the same scale. A smaller area of the red dashed box denotes the different languages' semantics are more tightly entangled. GL-CLEF's area is 35490 (182x195), while HC$^2$L's area is 31450 (170x185).}
 \label{fig: langcluster}
\end{figure}

\subsubsection{Multilingual Clustering}

We visualize the sentence embeddings of test samples in all languages, as shown in Fig. \ref{fig: langcluster}.
GL-CLEF can generally obtain promising representations as different languages' sentence embeddings overlap with each other.
However, in Fig. \ref{fig: langcluster}, HC$^2$L's red dashed box area is around 12\% smaller than GL-CLEF's, indicating that our HC$^2$L generates more tightly entangled sentence embeddings, 
proving that our model can generate more consistent semantic representations for different languages.
This can be attributed to the fact that our model performs explicit label-aware semantics knowledge transfer and label-aware multilingual semantics alignments.

\subsection{Parameter Analysis}

$\beta_I$, $\beta_S$ and $\beta_J$ control the balance of the single-task supervised CL and joint-task supervised CL.
Besides, they determine the value of $\mathcal{L}_{\text{slscl}}$.
$\gamma_1$ and $\gamma_2$ control the extent of cross-lingual supervised CL and multilingual supervised CL.
From Fig. \ref{fig: lambda} and \ref{fig: beta}, we can find that  on each hyper-parameter, the performances first grow and then drop after the peaks.
And the best values of $\beta_I$, $\beta_S$, $\beta_J$, $\gamma_1$ and $\gamma_2$ are 1e-2, 1e-4, 1e-4, 0.1 and 0.1, respectively.
Smaller values lead to inferior performances because they make the corresponding supervise signal weak, while larger values also make the performance drop because they harm the balance and dilute other supervise signals.

\begin{table}[t]
\centering
\fontsize{8}{10}\selectfont
\caption{Comparison on computation efficiency.}
\setlength{\tabcolsep}{1.6mm}{
\begin{tabular}{c|c|c}
\toprule
Models & \begin{tabular}[c]{@{}l@{}}GPU  Memory \\Required\end{tabular}  & \begin{tabular}[c]{@{}l@{}}Training Time\\ per Epoch\end{tabular}   \\ \midrule
GL-CLEF          & 20.7GB & 33s \\ \midrule
HC$^2$L (ours)    & 18.9GB & 35s \\
\bottomrule
\end{tabular}}
\label{table: computing_efficient}
\end{table}

\subsection{Computation Efficiency}

We compare our HC$^2$L with the state-of-the-art model GL-CLEF on computation efficiency, as shown in Table \ref{table: computing_efficient}.
Compared with GL-CLEF, our model can decrease the required GPU memory by 8.7\%, while costing more 6.1\% training time.
The GPU memory is dominated by the foundation LLM. The reason why our model occupies slightly less GPU memory is that the queue size of our model is 16, while the queue size of GL- CLEF is 32.
For inference, the two models cost the same time because they are based on the same SLU backbone and only perform contrastive learning in the training stage.



\section{Conclusion}\label{sec: conclusion}
This paper proposes Hybrid and Cooperative Contrastive Learning (HC$^2$L) for zero-shot cross-lingual SLU.
Apart from cross-lingual unsupervised CL, which has been exploited by the state-of-the-art model, we propose source language supervised CL, cross-lingual supervised CL and multilingual supervised CL to perform explicit label-aware semantics alignments.
Experiments on 9 languages verify the effectiveness of our model, whose four kinds of CL can cooperate to learn better semantic representations.

\ifCLASSOPTIONcompsoc
  \section*{Acknowledgments}
\else
  \section*{Acknowledgment}
\fi

This work was supported by National Key Research and Development Project of China (No. 2022YFC3502303) and Australian Research Council Grant (No.DP200101328).
Ivor W. Tsang was also supported by A$^*$STAR Centre for Frontier AI Research.

\ifCLASSOPTIONcaptionsoff
  \newpage
\fi

\normalem
\bibliographystyle{IEEEtran}
\bibliography{ref.bib}
%
%
\begin{IEEEbiography}[{\includegraphics[width=1in,height=1.25in,clip,keepaspectratio]{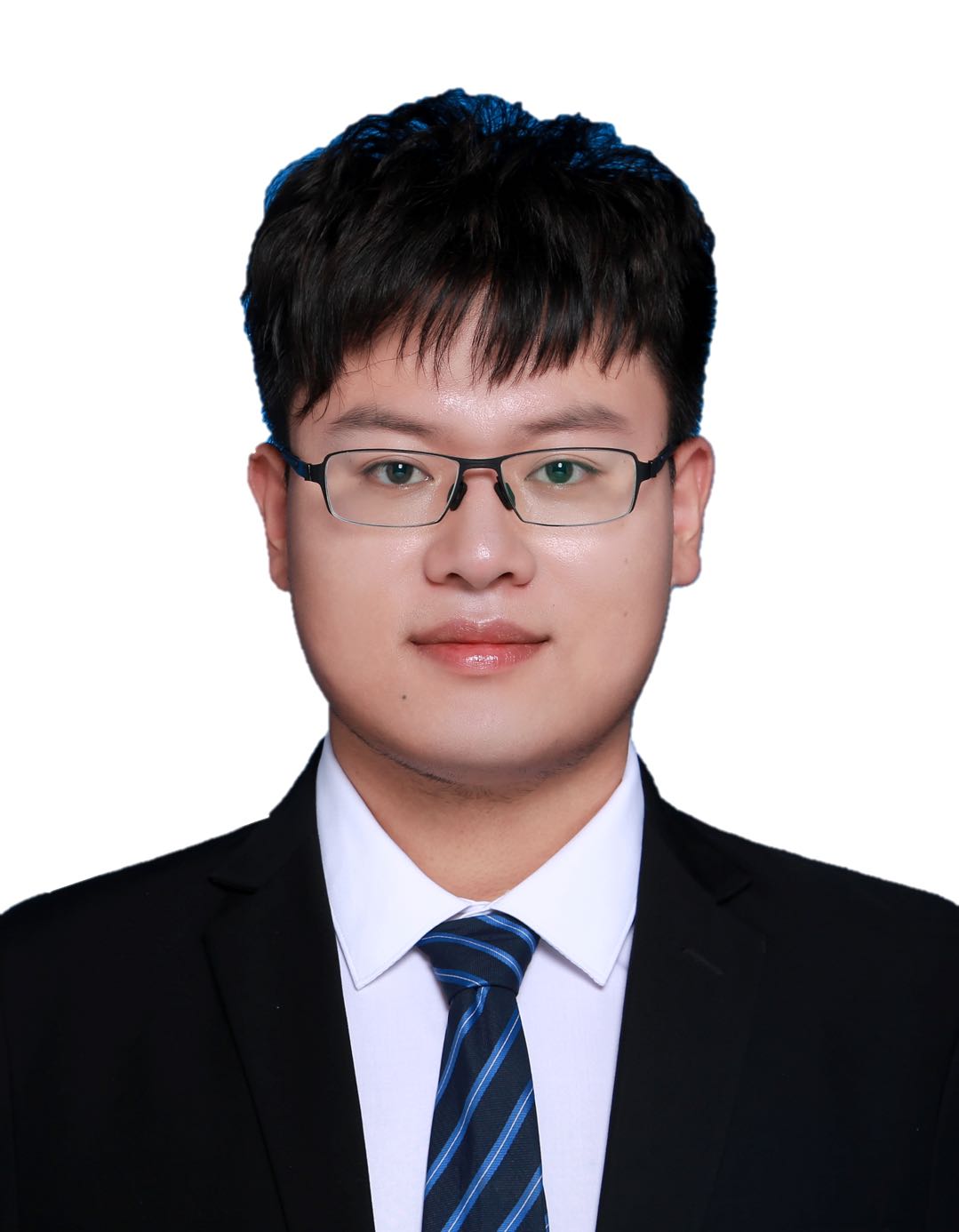}}]{Bowen Xing}
is an associate professor at school of computer and communication engineering, University of Science and Technology Beijing.
He received his B.E. degree and Master degree in computer science from Beijing Institute of Technology in 2017 and 2020, respectively. He obtained his PhD degree in artificial intelligence in 2024, from Australian Artificial Intelligence Institute (AAII), University of Technology Sydney (UTS), under the supervision of Professor Ivor W. Tsang.
His research focuses on natural language processing.
\end{IEEEbiography}

\begin{IEEEbiography}[{\includegraphics[width=1in,height=1.25in,clip,keepaspectratio]{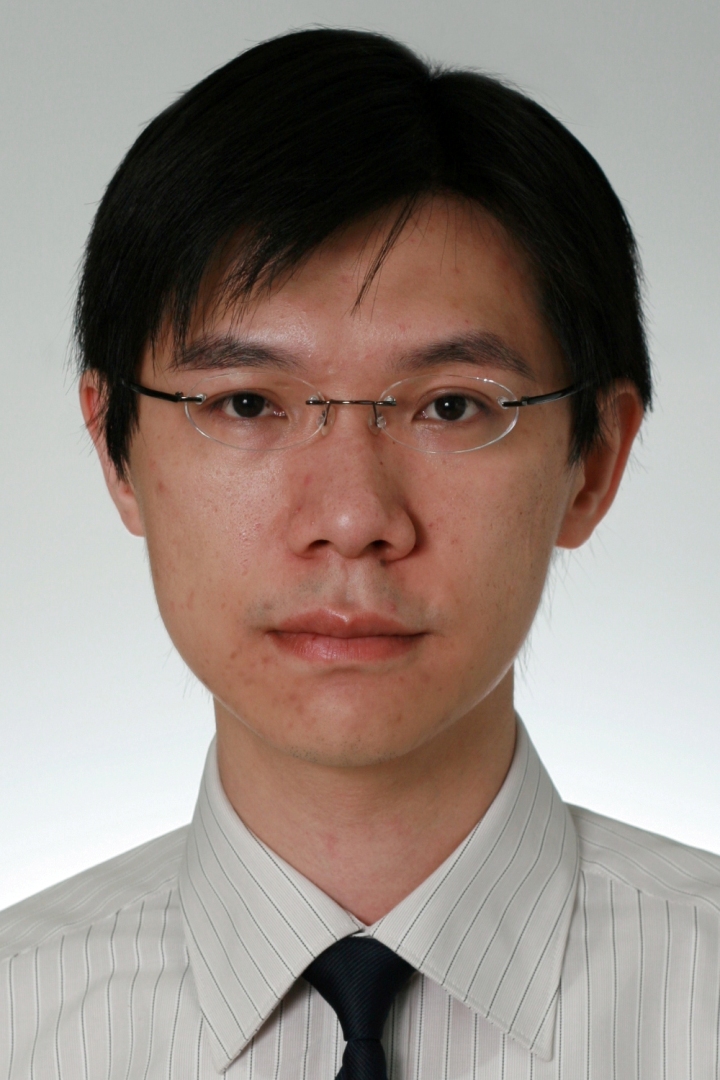}}]{Ivor W. Tsang} is an IEEE Fellow and the Director of A*STAR Centre for Frontier AI Research (CFAR). Previously, he was a Professor of Artificial Intelligence, at University of Technology Sydney (UTS), and Research Director of the Australian Artificial Intelligence Institute (AAII).
His research focuses on transfer learning, deep generative models, learning with weakly supervision, big data analytics for data with extremely high dimensions in features, samples and labels. His work is recognized internationally for its outstanding contributions to those fields.
In 2013, Prof Tsang received his ARC Future Fellowship for his outstanding research on big data analytics and large-scale machine learning. 
In 2019, his JMLR paper ``Towards ultrahigh dimensional feature selection for big data'' received the International Consortium of Chinese Mathematicians Best Paper Award. In 2020, he was recognized as the AI 2000 AAAI/IJCAI Most Influential Scholar in Australia for his outstanding contributions to the field, between 2009 and 2019. His research on transfer learning was awarded the Best Student Paper Award at CVPR 2010 and the 2014 IEEE TMM Prize Paper Award. In addition, he received the IEEE TNN Outstanding 2004 Paper Award in 2007 for his innovative work on solving the inverse problem of non-linear representations. Recently, Prof Tsang was conferred the IEEE Fellow for his outstanding contributions to large-scale machine learning and transfer learning.
Prof Tsang serves as the Editorial Board for the JMLR, MLJ, JAIR, IEEE TPAMI, IEEE TAI, IEEE TBD, and IEEE TETCI. He serves as a Senior Area Chair/Area Chair for NeurIPS, ICML, AAAI and IJCAI, and the steering committee of ACML.
\end{IEEEbiography}

\end{document}